\documentclass[10pt,twocolumn,letterpaper]{article}

\usepackage{cvpr}              %

\definecolor{cvprblue}{rgb}{0.21,0.49,0.74}
\usepackage[pagebackref,breaklinks,colorlinks,allcolors=cvprblue]{hyperref}

\def\paperID{xxxxx} %
\def\confName{CVPR}
\def\confYear{2026}

\def\eg{\emph{e.g}\onedot} 
\def\ie{\emph{i.e}\onedot} 
 
 \def\vs{\emph{vs}\onedot}

\newcommand{\tabref}[1]{Tab.~\ref{#1}}

\newcommand{\secref}[1]{Sec.~\ref{#1}}

\newlength\savewidth\newcommand\shline{\noalign{\global\savewidth\arrayrulewidth
  \global\arrayrulewidth 1pt}\hline\noalign{\global\arrayrulewidth\savewidth}}
\newcommand{\tablestyle}[2]{\setlength{\tabcolsep}{#1}\renewcommand{\arraystretch}{#2}\centering\footnotesize}

\usepackage{multirow}
\usepackage{xcolor}
\usepackage{colortbl}
\definecolor{baselinecolor}{gray}{.9}
\newcommand{\baseline}[1]{\cellcolor{baselinecolor}{#1}}

\newcommand{\modelname}{FreqFlow\xspace}

\title{Frequency-Aware Flow Matching for High-Quality Image Generation}
\author{Sucheng Ren$^1$\quad Qihang Yu$^2$\quad Ju He$^2$\quad Xiaohui Shen$^2$ \quad Alan Yuille$^1$ \quad Liang-Chieh Chen$^2$ \\
$^1$Johns Hopkins University \quad $^2$ByteDance 
}

\begin{document}
\maketitle
\begin{abstract}
    Flow matching models have emerged as a powerful framework for realistic image generation by learning to reverse a corruption process that progressively adds Gaussian noise. However, because noise is injected in the latent domain, its impact on different frequency components is non-uniform. As a result, during inference, flow matching models tend to generate low-frequency components (global structure) in the early stages, while high-frequency components (fine details) emerge only later in the reverse process.
    Building on this insight, we propose Frequency-Aware Flow Matching (\modelname), a novel approach that explicitly incorporates frequency-aware conditioning into the flow matching framework via time-dependent adaptive weighting. We introduce a two-branch architecture: (1) a frequency branch that separately processes low- and high-frequency components to capture global structure and refine textures and edges, and (2) a spatial branch that synthesizes images in the latent domain, guided by the frequency branch's output.
    By explicitly integrating frequency information into the generation process, \modelname ensures that both large-scale coherence and fine-grained details are effectively modeled—low-frequency conditioning reinforces global structure, while high-frequency conditioning enhances texture fidelity and detail sharpness.
    On the class-conditional ImageNet-256 generation benchmark, our method achieves state-of-the-art performance with an FID of 1.38, surpassing the prior diffusion model DiT and flow matching model SiT by 0.79 and 0.58 FID, respectively. Code is available at \url{https://github.com/OliverRensu/FreqFlow}.
\end{abstract}

\section{Introduction}
Recent advancements in generative modeling have fueled significant progress in image synthesis, driven by breakthroughs in diffusion-based~\cite{song2020score,diff1,diff2,diff3} and flow-matching methods~\cite{liu2022flow,lipman2022flow,shin2025deeply,ren2025beyond}. Among these approaches, flow matching models have emerged as a powerful framework for generating high-quality images by learning to reverse a noise corruption process. Specifically, these models sample a continuous-time trajectory between the data distribution and a simple Gaussian prior by aligning probability flows, enabling stable training dynamics and strong performance on large-scale image generation tasks.

\begin{figure}
    \centering
    \includegraphics[width=\linewidth]{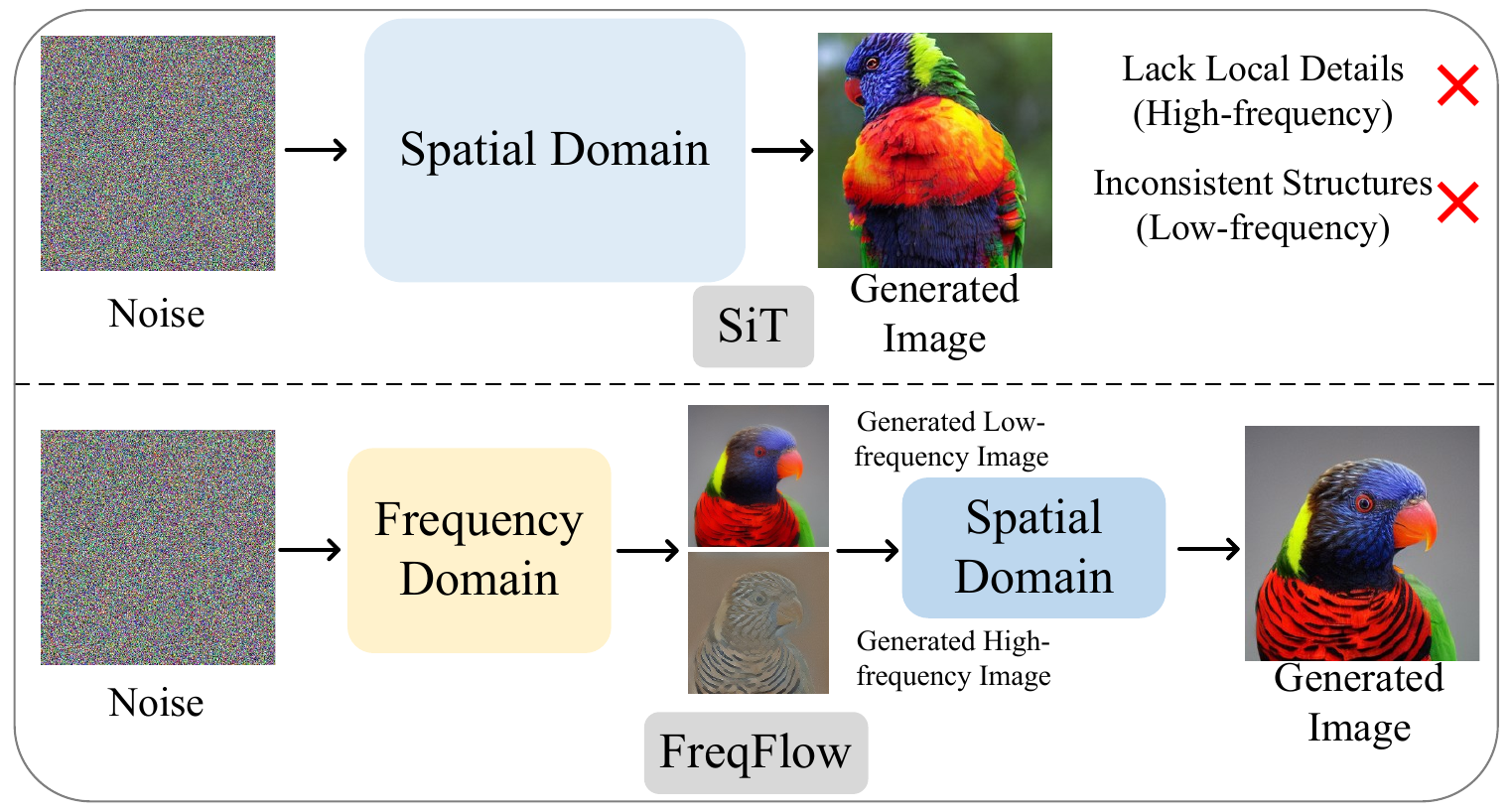}
    \caption{\textbf{Flow matching in the spatial domain \vs \ frequency-aware flow matching.} Unlike previous flow matching models such as SiT~\cite{sit}, which operate purely in the spatial domain, our \modelname explicitly incorporates frequency information into the spatial branch. This enhances local detail refinement while preserving structural consistency, leading to improved image quality.
    }
    \label{fig:teaser2}
\end{figure}

Despite these advantages, existing flow matching methods~\cite{sd3,sit} inject noise uniformly across the spatial domain, leading to suboptimal preservation of frequency components and ultimately affecting the quality of generated images.
When noise is introduced directly to pixels~\cite{diff1} or latent representations~\cite{vae,ldm}, as in diffusion and flow matching processes, it propagates unevenly across the frequency spectrum. Empirically, these models reconstruct global structures in the early stages of the reverse process—capturing low-frequency information such as overall shapes and color distributions—while high-frequency details, including textures and edges, are refined later. However, without explicit guidance on how to prioritize different frequency components, flow matching models often produce slightly blurred or smoothed results, particularly in fine details.
This phenomenon reveals a fundamental gap: while flow matching models operate in the spatial domain, the corruption and recovery processes inherently affect different frequency components in a non-uniform manner—yet these frequency-domain characteristics are neither explicitly modeled nor effectively leveraged.

To address these limitations, we introduce Frequency-Aware Flow Matching (\modelname), a novel approach that integrates an adaptive frequency-conditioning mechanism into the flow matching framework (\cref{fig:teaser2}). Our method is motivated by the observation that high- and low-frequency components require different levels of emphasis at various stages of the reverse process.
\modelname employs a two-branch architecture: (1) a frequency branch that generates the low-frequency global structure and high-frequency refinements, and (2) a spatial branch that synthesizes images in the latent domain, guided by explicit frequency conditioning. A time-dependent adaptive weighting mechanism dynamically balances the contributions of both branches throughout the flow matching trajectory, ensuring that the model effectively captures both large-scale structure and fine-grained details.

During the early stages of generation, the frequency branch prioritizes low-frequency components, allowing the model to establish the overall layout and shape of the image. As the process progresses, high-frequency details are introduced, refining textures and edges. By aligning the generation process with the natural order of human perception—where we first recognize coarse structures before noticing fine details—\modelname enhances image fidelity and accelerates training convergence.

\begin{figure}[t!]
    \centering
    \includegraphics[width=0.8\linewidth]{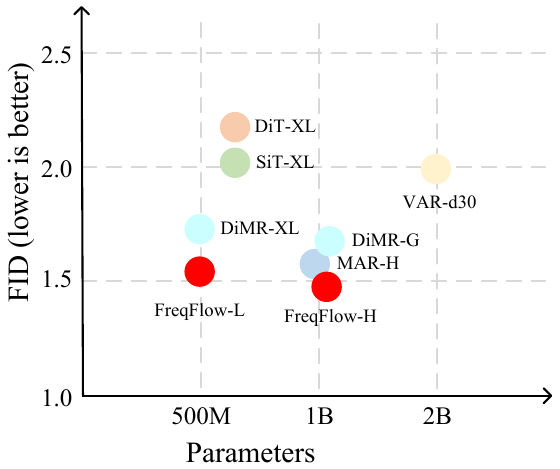}
    \vspace{-2mm}
    \caption{
    \textbf{Parameters \vs \ FID.} Our \modelname-L outperforms DiT-XL~\cite{dit} and SiT-XL~\cite{sit} by 0.73 and 0.52 FID, respectively, while using fewer parameters. Under comparable parameter budgets, \modelname-H surpasses DiMR-G~\cite{liu2024alleviating} and MAR-H~\cite{mar} by 0.15 and 0.07 FID, demonstrating superior efficiency and performance.
    }
    \label{fig:teaser}
\end{figure}

Experimental results on standard ImageNet-256 generation benchmark~\cite{deng2009imagenet} demonstrate that \modelname achieves an FID of 1.38, outperforming the state-of-the-art diffusion model DiT~\cite{dit} and flow matching model SiT~\cite{sit} by 0.79 and 0.58 FID, respectively (\cref{fig:teaser}).

\section{Related Work}
\noindent\textbf{Diffusion- and Flow- based Model.}
Recent advancements in image generation have been driven by diffusion models, which surpass traditional frameworks like Generative Adversarial Networks (GANs)~\cite{gan} through iterative diffusion and denoising processes~\cite{ldm, dit, mar, diff1, diff2, diff3, liu2024alleviating,ren2025grouping, jj1, jj2}. A key breakthrough is the Latent Diffusion Model (LDM)~\cite{ldm}, which shifts the diffusion process from pixel space to latent representations~\cite{vae}, significantly improving computational efficiency. This transition enables high-resolution image generation with reduced resource demands. Expanding on this foundation, DiT~\cite{dit} and U-ViT~\cite{uvit} integrate Transformer-based architectures~\cite{vaswani2017attention, dosovitskiy2020image,ren1,ren2,ren3,ren4} into the latent space, replacing conventional convolutional U-Nets~\cite{unet} and further enhancing performance in image synthesis.

In parallel, flow matching models~\cite{lipman2022flow, liu2022flow, albergo2022building, sd3, black2024flux1.1, yang20241, he2025flowtok} redefine the forward diffusion process by directly mapping data distributions to a standard Gaussian, streamlining the transformation from noise to structured data. This approach provides a more direct and computationally efficient alternative to traditional diffusion models. 
Notably, SiT~\cite{sit} extends this innovation by integrating DiT~\cite{dit} with flow matching, improving efficiency by establishing a more direct correspondence between distributions.
Unlike these methods, the proposed \modelname explicitly incorporates frequency-aware generation, leveraging a dedicated frequency branch to separately model low- and high-frequency components. This enhances fine-grained textures while preserving structural information through integration with the spatial branch, improving both image coherence and detail.

\noindent\textbf{Frequency in Image Generation.}
Several works have explored the role of frequency in image generation. Katja \etal~\cite{ganfreq} analyze high-frequency artifacts in GANs, highlighting their impact on synthesis quality. FreeU~\cite{freeu} improves U-Net-based denoising by re-weighting skip connections and backbone feature maps, balancing high-frequency detail preservation with semantic denoising, thereby enhancing generation quality without fine-tuning.
FouriScale~\cite{fouriscale} introduces a training-free frequency-domain approach that modifies pre-trained diffusion models using atrous convolutions~\cite{chen2015semantic,chen2017deeplab} and low-pass filtering to address resolution-based challenges, enabling high-resolution, structurally consistent image generation across various aspect ratios.
In the context of video generation, FRAG~\cite{frag} enhances video editing by incorporating the Frequency Adapting Group, which preserves high-frequency details during denoising. This prevents blurring and flickering, improving both consistency and fidelity in the final output.
Similarly, \modelname explicitly incorporates frequency-aware processing by leveraging a dedicated frequency branch to generate the low-frequency global structure and refine high-frequency details, ensuring improved coherence and sharpness in image synthesis.

\section{Method}
In this section, we first introduce the fundamentals of flow matching in~\secref{sec:flow_matching}.
Next, we analyze flow matching from a frequency perspective in~\secref{sec:revisit}.
Finally, we present our proposed \modelname in~\secref{sec:our_method}.

\subsection{Preliminaries: Flow Matching}
\label{sec:flow_matching}

Flow matching~\cite{liu2022flow,lipman2022flow} is a generative modeling framework that learns a continuous transformation between a simple initial distribution (typically Gaussian noise) and a complex target data distribution. Unlike diffusion models~\cite{diff2,ldm}, 
which reverse a stochastic noise corruption process, flow matching constructs a deterministic flow that smoothly transports samples from the source to the target distribution over a continuous time horizon.

During training, given an image (or its latent) $X$ from the data distribution, a flow matching model randomly samples a time step $t \in [0,1]$ and a noise sample $N \sim \mathcal{N}(0, I)$ from the source distribution.
The intermediate latent representation $X_t$ is then constructed as:
\begin{equation}
    X_t = (1 - t)\cdot X + t \cdot N.
\end{equation}
The goal is to estimate the velocity field $V_t$, which describes the direction from the source to the target distribution. Taking the derivative of $X_t$ with respect to $t$, we obtain:
\begin{equation}
\begin{split}
     V_t &= \dfrac{dX_t}{dt} \\
            &= N - X,
\end{split}
\end{equation}
where $V_t$ represents the ideal velocity that steers the intermediate distribution toward the target data distribution. To learn this velocity field, the model minimizes a flow matching loss, which measures the discrepancy between the model's predicted velocity and the true velocity field at each time step:
\begin{equation}
    \mathcal{L} =  \left\| f_\theta\left(X_t, t\right) - V_t \right\|^2,
\end{equation}
where $f_\theta$ is the model parameterized by $\theta$.
By optimizing this objective, flow matching ensures that the learned vector field accurately transports samples from noise distribution to data distribution in a continuous and stable manner.

\begin{figure}
    \centering
    \includegraphics[width=0.8\linewidth]{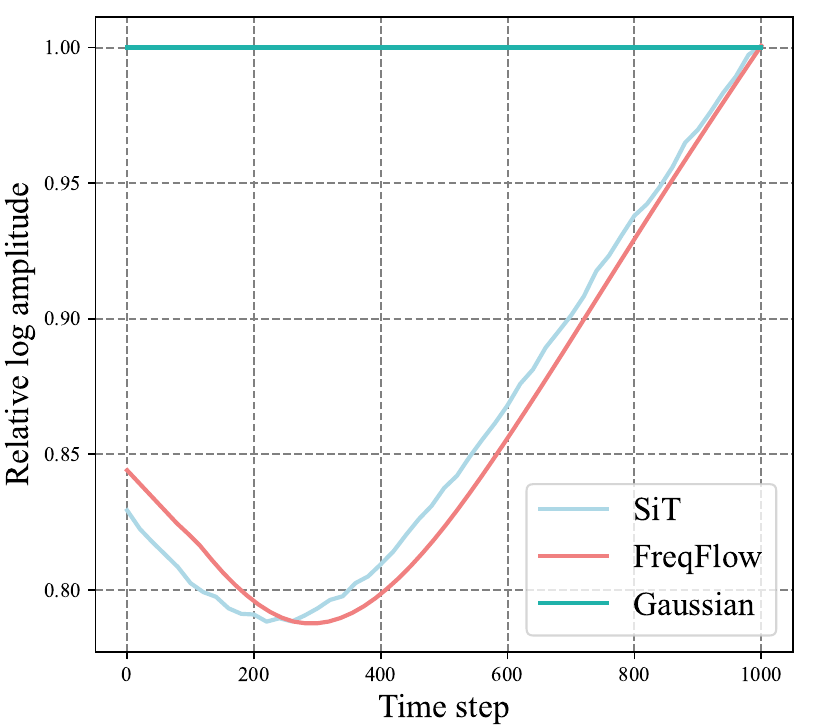}
    \caption{\textbf{Relative log amplitudes of frequency cross time steps from 1000 (pure Gaussian noise) to 0 (clean image).}
    Flow Matching models introduce low-frequency components in the early stages and high-frequency components in the later stages of the reverse process. Compared to SiT~\cite{sit}, our \modelname constructs global structures (low-frequency information) more efficiently—reaching the lowest log amplitude earlier in the process (time step 200 \vs \ 280). Additionally, \modelname progressively refines fine details (high-frequency information) in the final stages, resulting in a higher relative log amplitude at time step 0.
    }
    \label{fig:freq}
\end{figure}

\subsection{Flow Matching from a Frequency Perspective}
\label{sec:revisit}

To analyze how frequency components evolve throughout the flow matching process, we examine the log amplitude statistics of a pretrained flow matching model, SiT~\cite{sit}, as illustrated in ~\cref{fig:freq}. Our observations reveal that the model first generates low-frequency structural components from Gaussian noise (with a relative log amplitude 1), gradually incorporating high-frequency details in later stages. However, due to the lack of an explicit mechanism to prioritize different frequency bands, the generated images often exhibit slight blurring or smoothing, particularly in fine details.
This highlights a key limitation: while flow matching models effectively reconstruct images in the spatial domain, they do not explicitly account for the non-uniform corruption and recovery process in the frequency domain.

To quantify this effect, we measure the frequency error:
\begin{equation} \mathbb{E}[|\mathcal{F}_{real}|] - \mathbb{E}[|\mathcal{F}_{gen}|], \end{equation}
where $\mathcal{F}_{real}$ and $\mathcal{F}_{gen}$   denote the Fourier Transform of real and generated samples, respectively. The expectation is taken over all samples and frequency components.

As shown in~\cref{tab:freqerror}, SiT produces lower errors in low-frequency components but exhibits significantly larger errors in high-frequency components, indicating that SiT struggles to recover fine details. In contrast, our proposed \modelname reduces errors across both low- and high-frequency components, demonstrating its ability to generate images with improved structural coherence and finer details. We detail the design of \modelname in the following section and how it explicitly incorporates frequency-aware conditioning to address these limitations.

\begin{table}[t!]
    \centering
    \scalebox{0.95}{
    \begin{tabular}{c|c|c}
        model & low-frequency error& high-frequency error \\
        \shline
        SiT~\cite{sit}  & 0.08& 0.69\\ 
        \modelname &0.06&0.48\\
    \end{tabular} 
    }
    \caption{
    \textbf{Frequency error.}
    SiT exhibits larger errors in the high-frequency components, highlighting its difficulty in synthesizing fine details. In contrast, the proposed \modelname achieves lower errors across both low- and high-frequency components, demonstrating its ability to generate images with improved structural coherence and finer details.
    }
    \label{tab:freqerror}
\end{table}

\begin{figure*}
    \centering
    \includegraphics[width=0.8\linewidth]{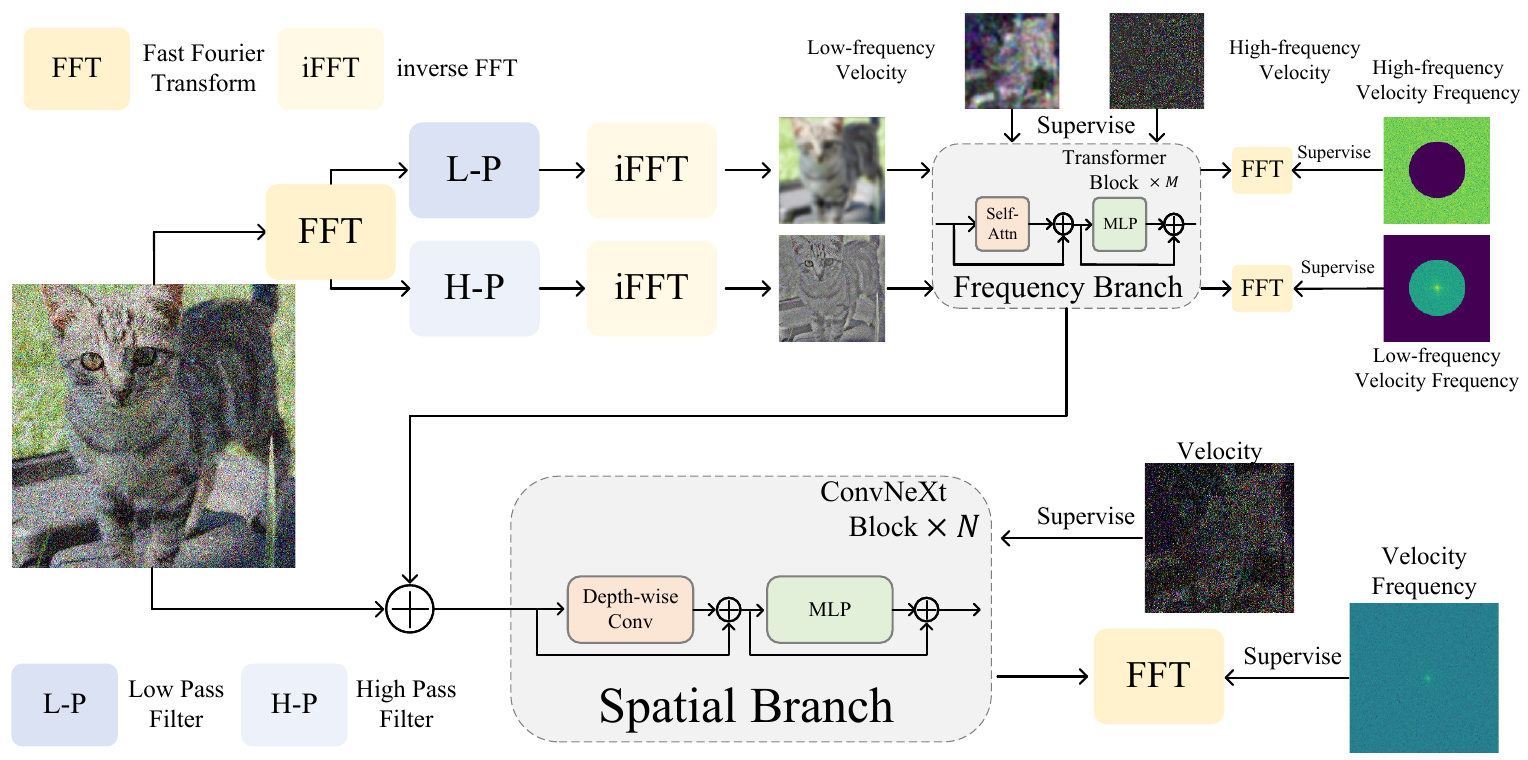}
    \caption{
    \textbf{Overview of \modelname.}
    \modelname features a two-branch design: (1) a frequency branch that captures the low-frequency global structure and high-frequency details (\eg, edges), and (2) a spatial branch that synthesizes images in the pixel or latent domain, guided by the frequency branch’s output.
    During training, the input noisy image is decomposed into low- and high-frequency components using low-pass and high-pass filters. The frequency branch processes these components with supervision from corresponding low- and high-frequency velocity fields. The spatial branch then refines the image while integrating frequency-aware features, supervised by both ordinary velocity and frequency velocity, ensuring improved structure and detail preservation. $\oplus$: Element-wise addition.
    }
    \label{fig:method}
\end{figure*}

\subsection{\modelname: A Frequency-aware Model}
\label{sec:our_method}

\noindent\textbf{Overview.} Unlike conventional flow matching models that implicitly generate frequency information, \modelname explicitly incorporates and manipulates different frequency components throughout the generation process. To achieve this, \modelname adopts a two-branch architecture. The frequency branch (\cref{sec:freq_branch}) separately processes low- and high-frequency components, capturing global structure while refining fine details. The spatial branch (\cref{sec:spatial_branch}) then synthesizes images in the latent domain (or directly in the pixel domain for lower resolutions), guided by the frequency branch’s output.
Given the nature of this two-branch structure, the model is trained with dual-domain supervision (\cref{sec:multi_supervision}), ensuring effective frequency-aware generation across different scales.
\cref{fig:method} presents an overview of \modelname, and the following subsections detail its design.

\subsubsection{Frequency Branch}
\label{sec:freq_branch}

Given a noisy image $X_t$ (or its noisy latent) of size \( H \times W \) at time step $t$, we first transform it into the frequency domain using the Discrete Fourier Transform (DFT):
\begin{equation}
F_t(u, v) = \sum_{x=0}^{H-1} \sum_{y=0}^{W-1} X_t(x, y) e^{-j2\pi\left( \frac{ux}{H} + \frac{vy}{W} \right)},
\label{eq:dft}
\end{equation}
where $X_t(x, y)$ represents the pixel value at position $(x, y)$, $F_t(u, v)$ is the corresponding complex-valued frequency component at $(u, v)$, and $j$ is the imaginary unit.

To explicitly process frequency components, we apply high-pass and low-pass Gaussian filters to separate high-frequency details ($H_t$) and low-frequency structural information ($L_t$):
\begin{equation}
    \begin{split}
        H_t(u, v) &= F_t(u, v) \cdot \left[1 - e^{-\frac{(u - \frac{H}{2})^2 + (v - \frac{W}{2})^2}{2\sigma_H^2}}\right], \label{eq:high_pass} \\
    L_t(u, v) &= F_t(u, v) \cdot e^{-\frac{(u - \frac{H}{2})^2 + (v - \frac{W}{2})^2}{2\sigma_L^2}}, 
    \end{split}
\end{equation}
where $\sigma_H$ and $\sigma_L$ control the cutoff frequencies for high-pass and low-pass filtering. The high-pass filter enhances edges, textures, and fine details, while the low-pass filter smooths the image and preserves overall structure.

After filtering, we reconstruct spatial representations via the inverse DFT (IDFT):
\begin{equation}
\begin{split}
    X_t^H(x, y) &= \frac{1}{HW} \sum_{u=0}^{H-1} \sum_{v=0}^{W-1} H_t(u, v) e^{j2\pi\left( \frac{ux}{H} + \frac{vy}{W} \right)},\\
    X_t^L(x, y) &= \frac{1}{HW} \sum_{u=0}^{H-1} \sum_{v=0}^{W-1} L_t(u, v) e^{j2\pi\left( \frac{ux}{H} + \frac{vy}{W} \right)}.
\end{split}
\label{eq:idft}
\end{equation}

To efficiently model frequency components, we design a frequency branch that separately processes low- and high-frequency representations:
\begin{equation}
\begin{split}
    \hat{V}_t^L, h_t^L &= f_{low}(X^L_t, t, c),\\
    \hat{V}_t^H, h_t^H &= f_{high}(X^H_t, t, c),
\end{split}
\label{eq:freq_branches}
\end{equation}
where $f_{low}$ and $f_{high}$ are networks designed to process low- and high-frequency components, respectively.
Here, $t$ represents the time step, $c$ is the class condition or other conditioning information, $\hat{V}_{t}^L$ and $\hat{V}_{t}^H$ are the predicted velocities for low and high frequency components, and $h_t^L$ and $h_t^H$ are the corresponding feature representations.

\noindent\textbf{Adaptive Frequency Integration.}
As discussed in~\cref{sec:revisit}, frequency components play different roles at different stages of the generation process: low frequencies dominate in early stages, while high frequencies refine details later. To incorporate this insight, we introduce an adaptive, time-dependent frequency integration mechanism:
\begin{equation}
\begin{split}
    \omega_t &= \sigma(\mathrm{MLP}(h_t^L, h_t^H, t)), \\
    h_t &= \omega_t \odot h_t^L + (1-\omega_t) \odot h_t^H,
\end{split}
\label{eq:adaptive_integration}
\end{equation}
where $\sigma$ is a sigmoid activation function, MLP is a multi-layer perceptron processing concatenated features and time step information, and $\odot$ denotes element-wise multiplication. The adaptive weight $\omega_t$ dynamically adjusts the contribution of low- and high-frequency components at different time steps.

\begin{figure}
    \centering
    \includegraphics[width=0.8\linewidth]{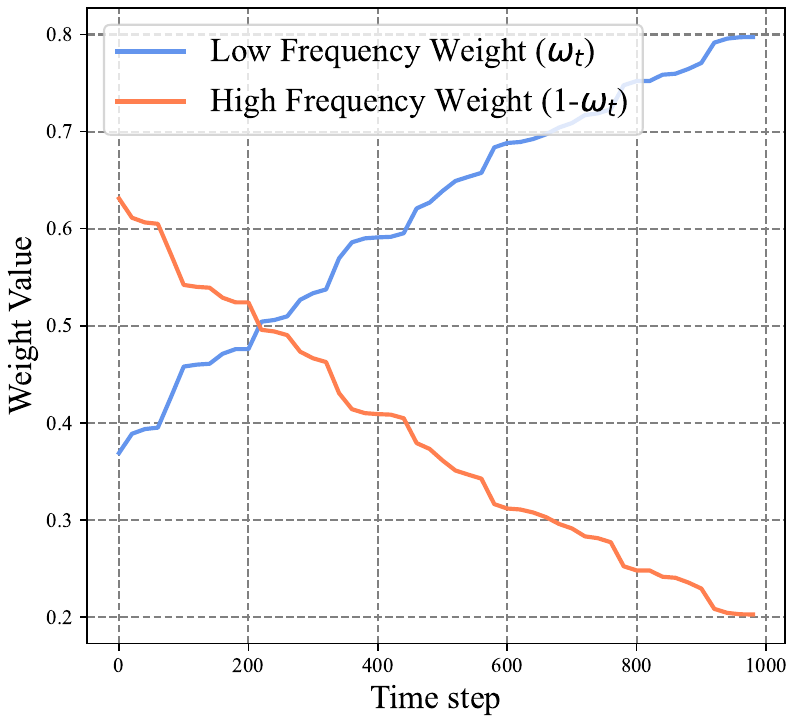}
    \caption{
    \textbf{Visualization of adaptive frequency integration during the reverse process from time step 1000 (pure Gaussian noise) to 0 (clean image).}
    The learned integration weights of low- ($\omega_t$) and high- ($1-\omega_t$) frequency components demonstrate that \modelname prioritizes low-frequency structure in the early stages (\ie, large time steps) and progressively shifts focus to high-frequency details toward the end (\ie, small time steps).
    }
    \label{fig:omega_viz}
\end{figure}

As shown in~\cref{fig:omega_viz}, we visualize the integration weight $\omega_t$ across different time steps.
During the reverse process, from pure Gaussian noise at time step 1000 to a clean image at time step 0,  $\omega_t$ (Low Frequency Weight) is larger in the early stages, indicating that the model prioritizes low-frequency components to establish the global structure. As generation progresses, it gradually shifts focus to high-frequency details, refining textures and edges. In the final stage, \modelname continues to integrate low-frequency information to maintain structural coherence while enhancing fine details. This behavior aligns with our analysis of how flow matching models naturally handle frequency components over time.

\noindent\textbf{Unified Architecture.} In practice, instead of maintaining two specialized networks $f_{low}$ and $f_{high}$, we implement a unified frequency branch $f_{freq}$ that processes both frequency components simultaneously:
\begin{equation}
    \hat{V}_t^H, \hat{V}_t^L, h_t = f_{freq}(X^H_t, X^L_t, t, c).
\label{eq:unified_freq}
\end{equation}
This design ensures efficient computation while preserving the benefits of frequency-aware processing. 
We instantiate $f_{freq}$ with a Vision Transformer~\cite{dosovitskiy2020image}, which excels in modeling long-range dependencies—particularly beneficial for frequency-based representations.

\subsubsection{Spatial Branch}
\label{sec:spatial_branch}

Given the noisy image $X_t$ (or its noisy latent) and the frequency branch output $h_t$ at time step $t$, we first combine them using a $merge$ operation, followed by processing with the spatial branch $f_{spatial}$:
\begin{equation}
    \hat{V}_{t} = f_{spatial}(merge(X_t, h_t), t, c),
\label{eq:spatial_branch}
\end{equation}
where $merge$ is implemented as element-wise addition (we ablate alternative design choices in the experiments), and $f_{spatial}$ is instantiated using  ConvNeXt~\cite{convnext}.
Compared to Vision Transformer (ViT)~\cite{dosovitskiy2020image}, ConvNeXt   is more effective at capturing high-frequency details, such as edges and textures. Additionally, the frequency branch output $h_t$ provides complementary low- and high-frequency information, ensuring the preservation of global structure while enhancing fine details in the final image synthesis.

\subsubsection{Training Strategy: Dual-domain Supervision}
\label{sec:multi_supervision}

To ensure effective learning across both spatial and frequency domains, we employ a dual-domain loss function. Specifically, in addition to the standard spatial domain loss $\mathcal{L}_{s}$, we introduce a frequency domain loss $\mathcal{L}_{f}$, which directly supervises the model output in the frequency domain:
\begin{equation}
\begin{split}
    \mathcal{L}_{s}(y, \hat{y}) &= \left\|y - \hat{y} \right\|_2^2,\\
    \mathcal{L}_{f}(y, \hat{y}) &= \left\|\text{FFT}(y) - \text{FFT}(\hat{y}) \right\|_2^2,
\end{split}
\label{eq:losses}
\end{equation}
where FFT represents the Fast Fourier Transform.

To comprehensively guide learning, our final training objective integrates supervision across different components and domains:
\begin{equation}
\begin{split}
    \mathcal{L} &= \mathcal{L}_{s}(\hat{V}_{t}, V_{t}) + \mathcal{L}_{f}(\hat{V}_{t}, V_{t}) \\
    &+ \alpha( \mathcal{L}_{s}(\hat{V}_{t}^H, V_{t}^H) + \mathcal{L}_{s}(\hat{V}_{t}^L, V_{t}^L) \\
    & \ \ \ \ + \mathcal{L}_{f}(\hat{V}_{t}^H, V_{t}^H) + \mathcal{L}_{f}(\hat{V}_{t}^L, V_{t}^L)),
\end{split}
\label{eq:total_loss}
\end{equation}
where $\alpha$ is a hyperparameter that balances the contribution of loss terms. 
In ~\cref{eq:total_loss}, the first two terms, applied to the spatial branch, provide global supervision in both the spatial and frequency domains. The remaining terms, applied to the frequency branch, explicitly supervise individual frequency components (low or high) within their respective domains.
This comprehensive loss function enables \modelname to effectively capture generation dynamics across different frequency bands, enhancing both image quality and diversity.
We set $\alpha=0.5$ by default, and our experiments show that performance remains consistent across different values of $\alpha$.

\noindent\textbf{Discussion.}
Thanks to its architecture and loss design, \modelname follows a frequency-aware generation trajectory: it first rapidly converges from Gaussian noise to low-frequency structural content, then progressively refines high-frequency details in later stages. As illustrated in~\cref{fig:freq}, this behavior is enabled by explicitly decoupling low-frequency modeling in early stages from high-frequency refinement in later stages.
The effectiveness of this approach is further quantified in~\cref{tab:freqerror}, where \modelname demonstrates lower low-frequency errors and significantly reduced high-frequency errors compared to SiT~\cite{sit}. This highlights its superior spectral alignment throughout the generation process, leading to sharper and more structurally coherent images.

\section{Experimental Results}
\label{sec:experiments}
In this section, we first describe the experimental setup (\cref{sec:setup}), followed by the main results (\cref{sec:main_results}).
We then conduct ablation studies on key design choices (~\cref{sec:ablation}).

\subsection{Experimental Setup}
\label{sec:setup}
We train \modelname for class-conditional image generation on ImageNet~\cite{deng2009imagenet} at resolutions 64$\times$64, 256$\times$256, and 512$\times$512. For 64$\times$64 images, we train \modelname directly in the pixel space. For 256$\times$256 and 512$\times$512 images, following prior works~\cite{dit,sit}, we utilize a pre-trained variational autoencoder (VAE) from Stable Diffusion~\cite{ldm} to extract latent representations of size 32$\times$32 and 64$\times$64, respectively. We then train \modelname to model these latent representations.
To evaluate image quality, we compute Fréchet Inception Distance (FID)~\cite{fid} on 50K generated samples. For fair comparisons, we follow the same evaluation protocol as the baselines and additionally report Inception Score (IS)~\cite{is} and Precision/Recall metrics.

\subsection{Main Results}
\label{sec:main_results}

\begin{table}[t!]
\centering
\begin{tabular}{l|c|cc}
model &  \#params.& FID$\downarrow$ & IS$\uparrow$\\
\shline
U-ViT-M/4~\cite{uvit} & 131M  & 5.85 & 33.71  \\
U-ViT-L/4~\cite{uvit} & 287M & 4.26 & 40.66  \\
DiMR-M/3R~\cite{liu2024alleviating}& 133M  & 3.65 & 42.41 \\
DiMR-L/3R~\cite{liu2024alleviating} & 284M & 2.21 & 55.73  \\
\hline
\modelname-B & 134M & 1.92	 &59.34 \\
\end{tabular}
\caption{\textbf{Class-conditional generation on
ImageNet $64\times64$.}
\label{tab:in1k64}
}
\end{table}

\noindent\textbf{ImageNet-64.}
In~\tabref{tab:in1k64}, we present a quantitative comparison of \modelname on class-conditional image generation at 64$\times$64 resolution on ImageNet~\cite{deng2009imagenet}. \modelname-B surpasses all competing methods in both FID and Inception Score (IS) while maintaining a comparable parameter budget. Specifically, \modelname-B achieves an FID of 1.92 and an IS of 59.34 with 134M parameters, outperforming DiMR-L/3R~\cite{liu2024alleviating} by 0.29 FID, despite using significantly fewer parameters (134M \vs 284M). This result highlights the efficiency of our frequency-aware design.

\noindent\textbf{ImageNet-256.} 
\tabref{tab:in1k256} compares \modelname against a variety of generative models. Focusing first on diffusion- and flow-based methods, \modelname-L (507M parameters) improves upon DiT-XL/2~\cite{dit} and SiT-XL/2~\cite{sit} (675M) by 0.73 and 0.52 FID, respectively. When scaled to 1.08B parameters (\modelname-H), our approach further reduces FID to 1.38, setting a new state-of-the-art among flow-based generators. Notably, \modelname-H also achieves significant gains over other generative paradigms, such as GANs, autoregressive (AR) models, and mask-prediction methods.

\tabref{tab:nocfg} presents our results \textit{without} classifier-free guidance~\cite{ho2022classifier}. Even at 507M parameters, \modelname-L outperforms the larger DiMR-G/2R~\cite{liu2024alleviating} (1.06B parameters) by 0.42 FID. Meanwhile, \modelname-H achieves an FID of 2.45, surpassing prior diffusion methods—such as DiT-XL~\cite{dit} and DiMR-G~\cite{liu2024alleviating}—by margins of 7.17 and 1.11 FID.

\begin{table*}[t!]
\centering
\vspace{-2mm}
\scalebox{0.88}{
\begin{tabular}{l|c|c|c|ccccc}
model & type & epochs & \#params. & FID$\downarrow$&IS$\uparrow$ &Precision$\uparrow$ &Recall$\uparrow$\\
\shline
BigGAN~\cite{biggan}& GAN& -& 112M &6.95 &224.5& 0.89 &0.38\\
GigaGAN~\cite{gigagan}& GAN& - &569M  &3.45 &225.5 &0.84& 0.61\\
\hline
VQGAN~\cite{vqgan}& AR& 100 &1.4B & 15.78& 74.3& -& -\\
RQTran~\cite{rq} &AR& 50& 3.8B & 7.55& 134.0& -& -\\
VQGAN-re~\cite{vqgan}&AR &100& 1.4B& 5.20& 280.3& -& -\\
ViTVQ~\cite{vit-vqgan}& AR& 100& 1.7B& 4.17& 175.1& -& -\\
RQTran-re~\cite{rq} & AR &50& 3.8B & 3.80 &323.7& -& - \\
ViTVQ-re~\cite{vit-vqgan}& AR &100 &1.7B & 3.04& 227.4 &-& -\\
RAR-L~\cite{yu2024randomized} & AR & 400 & 461M & 1.70 & 299.5 & 0.81 & 0.60\\
\hline
VAR-d30~\cite{var}& VAR& 350& 2.0B & 1.97& 334.7 &0.81& 0.61\\
MVAR-d30~\cite{mvar}& VAR& 350& 3.0B & 1.78 331.2 0.83 0.61\\
\hline
FlowAR-H~\cite{flowar} & FlowAR & 400 & 1.9B & 1.65 & 296.5 & 0.83 & 0.60\\
\hline
MAR-H~\cite{mar}& MAR&800& 943M& 1.55& 303.7 &0.81& 0.62\\
\hline
MaskGIT~\cite{maskgit}& Mask.& 300 &227M & 6.18 &182.1 &0.80 &0.51\\
MaskGIT-re~\cite{maskgit}& Mask.& 300 &227M &4.02 &355.6 &- &-\\
RCG~\cite{rcg} &Mask.& 200& 502M  &3.49& 215.5&-&-\\
TiTok-S-128~\cite{yu2024image}& Mask. &800 &287M &1.97& 281.8&-&-\\
MaskBit~\cite{weber2024maskbit} & Mask. & 1080 & 305M & 1.52 & 328.6 & - & - \\
\hline
CDM~\cite{cdm} &Diff.&2158 &- &4.88& 158.7 & -&- \\
ADM-U~\cite{adm} &Diff. & 396 & 608M   & 3.94&  215.8& 0.83 &0.53 \\
LDM-4 ~\cite{ldm}&Diff. & 166 & 400M  & 3.60&247.7&-&- \\
Simple-Diffusion~\cite{diff1}&Diff. &-& 2B& 2.44& 256.3&-&- \\
U-ViT-H/2 ~\cite{uvit} &Diff.& 400 & 501M   & 2.29 &259.1& 0.86 &0.56 \\
DiT-XL/2~\cite{dit}&Diff. & 1399 & 675M  & 2.27& 278.2 &0.83 &0.57\\
DiffiT~\cite{hatamizadeh2023diffit} &Diff.&400& 561M&1.73& 276.5 &0.80& 0.62 \\
DiMR-XL/2R~\cite{liu2024alleviating} &Diff.&800& 505M& 1.70 &289.0 &0.79 &0.63\\
DiMR-G/2R~\cite{liu2024alleviating}&Diff. & 800& 1.06B& 1.63 &292.5 &0.79& 0.63\\
MDTv2-XL/2~\cite{gao2023mdtv2}&Diff. &800& 676M& 1.58& 314.7 &0.79 &0.65\\
\hline
SiT-XL/2~\cite{sit} &Flow.& 1399 & 675M & 2.06 &270.3&0.82 &0.59\\
\hline
\modelname-L (ours)&Flow. & 800 & 507M  &1.54 &295.6&0.80&0.63\\
\modelname-H (ours)&Flow. & 800 & 1.08B  & 1.38 &298.5 &0.81&0.64\\
\end{tabular}
}
\caption{\textbf{Class-conditional image generation on
ImageNet $256\times256$.}
We report training epochs, number of parameters (\#params), and FID-50K \textbf{with} Classifier-Free Guidance (CFG).
``-re'': rejection sampling.
}
\label{tab:in1k256}
\end{table*}

\begin{table}[t!]
\centering
\vspace{-1mm}
\begin{tabular}{l|c|c}
model &  \#params.& FID~(w/o CFG)$\downarrow$\\
\shline
LDM-4~\cite{ldm} & 400M & 10.56 \\
DiT-XL/2~\cite{dit}&  675M  &9.62 \\
ADM-U~\cite{adm}&  608M & 7.49 \\
U-ViT-H/2~\cite{uvit}&  501M & 6.58 \\
DiMR-XL/2R~\cite{liu2024alleviating} & 505M & 4.50 \\
DiMR-G/2R~\cite{liu2024alleviating} & 1.06B  &3.56 \\
\hline
\modelname-L & 507M &3.12	  \\
\modelname-H & 1.08B & 2.45	 \\
\end{tabular}
\caption{\textbf{Class-conditional image generation on
ImageNet $256\times256$ \emph{without} classifier-free guidance.}
}
\label{tab:nocfg}
\end{table}

\noindent\textbf{ImageNet-512.}
\cref{tab:in1k512} summarizes our result on ImageNet 512$\times$512 generation benchmark.
\modelname-L achieves an FID of 2.02, outperforming DiT-XL/2 and U-ViT-H/4 by margins of 1.02 and 2.03, respectively. \modelname-L also attains a high Inception Score of 285.3, surpassing several previously reported baselines such as DiffiT.

\begin{table}[]
\centering
\begin{tabular}{l|c|cc}
model &  \#params.& FID$\downarrow$ & IS$\uparrow$\\
\shline
BigGAN~\cite{biggan}  & 158M & 8.43 & 177.9 \\
StyleGAN-XL~\cite{sauer2022stylegan} & -  & 2.41 & 267.8\\
\hline
MaskGIT~\cite{maskgit} & 227M  & 7.32 & 156.0 \\
MaskGIT-re~\cite{maskgit} & 227M  & 4.46 & 342.0  \\
VAR-${d}36$-s~\cite{var}  & 2.35B  & 2.63 & 303.2  \\
\hline
ADM-G~\cite{adm}& 422M  & 7.72 & 172.7  \\
U-ViT-L/4~\cite{uvit} & 287M  & 4.67 & 213.3 \\
U-ViT-H/4~\cite{uvit}  & 501M  & 4.05 & 263.8\\
ADM-G, ADM-U~\cite{adm}  & 731M & 3.85 & 221.7 \\
D\scriptsize{IFFU}\small{SSM}-XL~\cite{yan2023diffusion}  & 673M & 3.41 & 255.0 \\
DiT-XL/2~\cite{dit}  & 675M  & 3.04 & 240.8\\
DiMR-XL/3R~\cite{liu2024alleviating}   & 525M & 2.89 & 289.8  \\
DiffiT~\cite{hatamizadeh2023diffit} & 561M & 2.67 & 252.1  \\
\hline
\modelname-L & 507M & 2.02&285.3 \\
\end{tabular}
\caption{\textbf{Class-conditional  generation on
ImageNet $512\times512$.}
}
\label{tab:in1k512}
\end{table}

\begin{figure*}[t]
    \centering
    \setlength{\tabcolsep}{1pt}
    \begin{tabular}{cccccc}
    \includegraphics[width=0.16\linewidth]{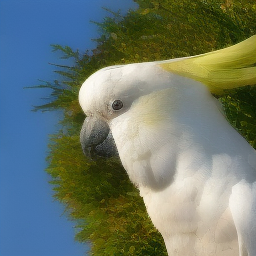}  & \includegraphics[width=0.16\linewidth]{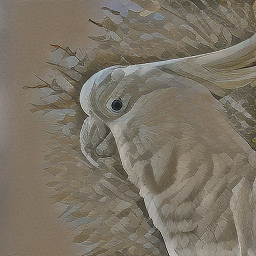} &\includegraphics[width=0.16\linewidth]{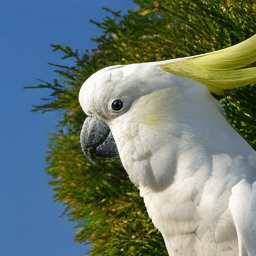}&
    \includegraphics[width=0.16\linewidth]{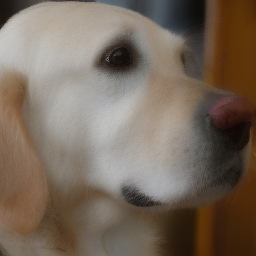}  & \includegraphics[width=0.16\linewidth]{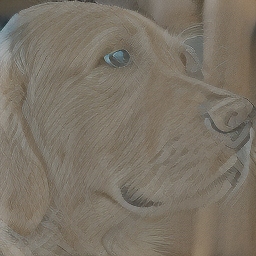} &\includegraphics[width=0.16\linewidth]{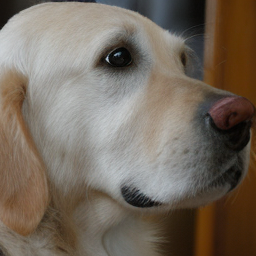}\\
      Low-frequency & High-frequency & Final & Low-frequency & High-frequency & Final \\
    \end{tabular}
    \caption{
    \textbf{Visualization of generated low-, high-frequency and final outputs.} The final output from the spatial branch is enhanced by the low- and high-frequency information provided by the frequency branch.
    }
    \label{fig:freq_vis}
\end{figure*}

\begin{figure*}[!t]
    \centering
    \vspace{-2mm}
    \includegraphics[width=\linewidth]{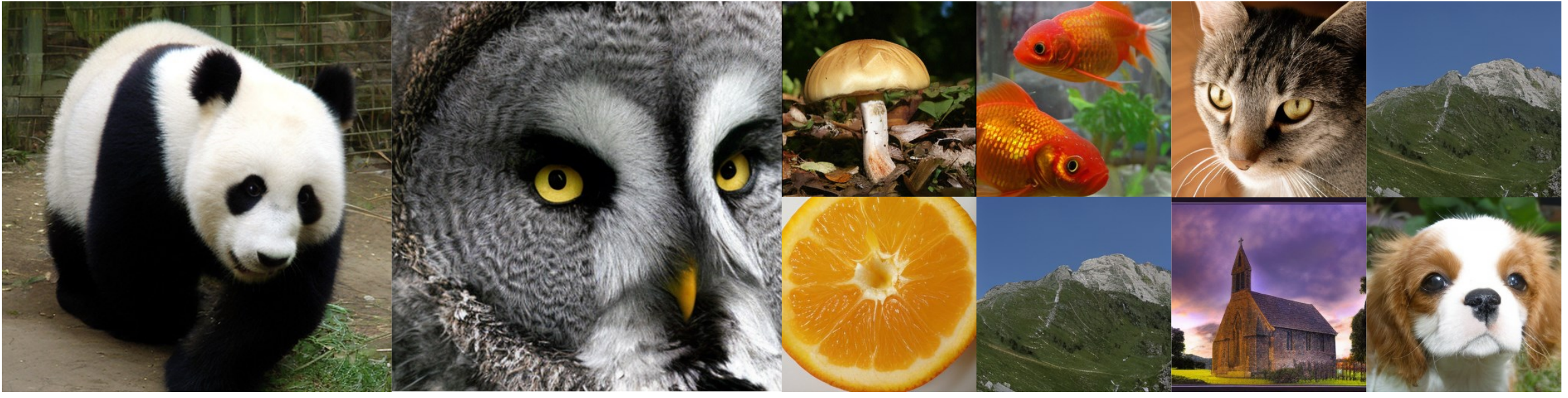}
    \caption{\textbf{Generations.}
    \modelname produces high-quality 512$\times$512 (1st and 2nd columns) and 256$\times$256 images (remaining columns).
    }
    \label{fig:visual}
\end{figure*}

\noindent\textbf{Qualitative Results.}
\cref{fig:freq_vis} and~\cref{fig:visual} visualize \modelname's generated samples, highlighting their quality and diversity.

\subsection{Ablation Studies}
\label{sec:ablation}

We perform ablation studies using \modelname-B (134M parameters) trained for 400 epochs on ImageNet-256, enabling efficient design iterations.

\noindent\textbf{Low \vs \ High Frequency.} We propose to explicitly introduce frequency information into flow matching models. As shown in~\cref{tab:component}, adding either low or high frequency component alone consistently improves performance over the baseline without any frequency information, demonstrating their effectiveness. However, the high-frequency component appears more influential overall, providing larger gains in both FID and IS. Notably, combining both low- and high-frequency information yields the best performance (FID = 2.95, IS = 231.5), underscoring the complementary benefits of leveraging all frequency components.
\begin{table}[]
    \centering
    \vspace{-1mm}
    \begin{tabular}{c|c|c|c}
        low-frequency& high-frequency& FID$\downarrow$ &IS$\uparrow$ \\
        \shline 
         & &3.86& 200.4\\
         $\checkmark$ & &3.55&208.5\\
            &$\checkmark$ &3.12&222.8\\
         \baseline{$\checkmark$}    &\baseline{$\checkmark$} &\baseline{2.95}&\baseline{231.5}\\
    \end{tabular}
    \caption{\textbf{Effectiveness of low- and high-frequency components.}
    Our final setting is labeled in gray.
    }
    \label{tab:component}
\end{table}

\noindent\textbf{Frequency Component Integration.} 
In~\cref{tab:fusion}, we ablate alternative design choices for the $merge$ operation in the spatial branch (\cref{eq:spatial_branch}): (1) \textit{cross attention}, where $h_t$ serves as the key and value while $X_t$ is the query; (2) \textit{channel concatenation}, which stacks $h_t$ and $X_t$ along the channel dimension; and (3) \textit{addition}, which combines them via element-wise summation. Among these, \textit{addition} achieves the best results (FID = 2.95, IS = 231.5), improving FID by 1.0 in FID over  \textit{cross attention}, and 0.51 over \textit{channel concatenation}. This highlights that a simple element-wise addition effectively integrates frequency cues with spatial features, enhancing image synthesis quality.

\begin{table}[]
    \centering
    \begin{tabular}{c|c|c}
        fusion scheme &  FID $\downarrow$& IS$\uparrow$\\
        \shline
        cross attention& 3.95&198.6\\
        channel concatenation & 3.46&224.8\\
        \baseline{addition}  & \baseline{2.95}&\baseline{231.5} \\
    \end{tabular}
    \vspace{-2mm}
     \caption{\textbf{Ablation on frequency component integration.}
    Our final setting is labeled in gray.
    }
    \label{tab:fusion}
\end{table}

\noindent\textbf{Loss at Frequency Branch.}
\modelname employs a two-branch design, where the frequency branch benefits from dedicated supervision. To assess its impact, we ablate the effect of frequency-based loss terms. As shown in~\cref{tab:loss}, incorporating these losses significantly enhances performance, reducing FID from 4.67 to 2.95 and increasing IS from 198.4 to 231.5. This targeted supervision helps the model refine fine details (high-frequency signals) while preserving overall structure (low-frequency signals).

\begin{table}[]
    \centering
    \begin{tabular}{c|c|c}
        loss on frequency branch & FID$\downarrow$& IS$\uparrow$ \\
        \shline
        &4.67& 198.4\\
        \baseline{$\checkmark$}  &\baseline{2.95}&\baseline{231.5}\\
    \end{tabular}
    \vspace{-2mm}
    \caption{\textbf{Ablation on the loss at the frequency branch.}
    Our final setting is labeled in gray.
    }
    \label{tab:loss}
\end{table}

\section{Conclusion}
We address a key limitation in flow matching models: uniform spatial noise injection neglects frequency characteristics, compromising high-frequency details. To overcome this, we propose \modelname, a frequency-aware framework with a two-branch architecture that separates low-frequency structure from high-frequency refinement. With time-dependent adaptive weighting, \modelname aligns generation with human visual perception—coarse structures first, fine details later. Our results underscore the importance of frequency-aware modeling for improved synthesis.

\subsection*{Acknowledgement}
This work is supported by ONR N000142412696.
{
    \small
    \bibliographystyle{ieeenat_fullname}
    \bibliography{main}
}
\clearpage
\setcounter{page}{1}

\renewcommand{\thesection}{\Alph{section}}
\setcounter{section}{0}

\section*{Appendix}
\label{sec:appendix}

The supplementary material includes the following additional information:

\begin{itemize}
\item \secref{sec:sup_variant} details the model variants of \modelname.
    \item \secref{sec:sup_hyper} details the hyper-parameters for \modelname.
    \item \secref{sec:sup_ablation} provides additional ablation studies.
    \item \secref{sec:limitation} includes the limitations and discussion.
    \item \secref{sec:freq_vis} presents visualizations of low- and high-frequency components (from the frequency branch) and final outputs (from the spatial branch) generated by \modelname.
    \item \secref{sec:sup_vis} provides additional visualization samples from the spatial branch of \modelname.
\end{itemize}

\section{Model Variants}
\label{sec:sup_variant}
\cref{tab:variant} presents the architectural details of our  model variants: \modelname-B, \modelname-L, and \modelname-H.

\begin{table}[h]
    \centering
    \begin{tabular}{c|c|c|c}
       model  &  depth & hidden size & \#params\\
       \hline
      \modelname-B   & (15, 12)&(768, 384)& 134M\\
       \modelname-L   & (39, 20)&(960, 480)& 507M\\
       \modelname-H   &  (57, 29)& (1152, 576) &1.08B\\
    \end{tabular}
    \caption{\textbf{\modelname model variants.} We provide detailed model configurations, including the depth and hidden size of the frequency and spatial branches. The first number in parentheses represents the design of the frequency branch, while the second corresponds to the spatial branch.
    }
    \label{tab:variant}
\end{table}

\section{Hyper-parameters for \modelname}
\label{sec:sup_hyper}
We detail the hyper-parameters of \modelname in~\tabref{tab:hparams}.

\begin{table}[h!]
\centering

\tablestyle{5.0pt}{1.1}
\begin{tabular}{l|c}
config \quad\quad\quad\quad\quad\quad\quad\quad & value \\
\hline
optimizer & AdamW~\cite{kingma2015adam,loshchilov2017decoupled} \\
optimizer momentum & (0.99, 0.99) \\
weight decay & 0.03 \\
batch size & 2048 \\
learning rate schedule & constant \\
peak learning rate & 2e-4 \\
total epochs & 800 \\
warmup epochs & 5\\
class label dropout rate & 0.1 \\
$\sigma_L$ & 8 \\
$\sigma_H$ &  2\\           
inference mode & ODE \\
inference steps & 250 \\
\end{tabular}
\caption{\textbf{Detailed Hyper-parameters of \modelname Models.}
}
\label{tab:hparams}
\end{table}
\section{Additional Ablation Studies}
\label{sec:sup_ablation}
In \modelname's frequency branch, we adopt a unified frequency branch $f_{freq}$ that processes both low- and high-frequency components simultaneously.
An alternative design, as discussed in Equation (8) of the main paper, involves using separate networks $f_{low}$ and $f_{high}$ to handle low- and high-frequency components individually.
In~\tabref{tab:branch}, we ablate this design choice, showing that the unified frequency branch improves FID by 0.49 compared to the separate architecture, demonstrating its effectiveness.

\begin{table}[ht!]
    \centering
    \begin{tabular}{c|c|c}
        frequency branch design &  FID $\downarrow$ & IS$\uparrow$\\
         \hline
        separate $f_{low}$ and $f_{high}$ & 3.44 & 210.2 \\
        \baseline{unified $f_{freq}$} & \baseline{2.95}  & \baseline{231.5}  \\
    \end{tabular}
    \caption{\textbf{Ablation on frequency branch architecture. } Our final setting is labeled in gray.    
    }
    \label{tab:branch}
\end{table}

\section{Limitations and Discussion}
\label{sec:limitation}
While our proposed \modelname achieves notable improvements over existing approaches on ImageNet, the largest \modelname-H only has about 1B parameters due to constraints on computational resources. We leave further scaling \modelname as future work.

\section{Visualization of Generated Low-/High-frequency and Final Outputs}
\label{sec:freq_vis}
\begin{figure*}[t]
    \centering
    \setlength{\tabcolsep}{3pt}
    \begin{tabular}{ccc}
    \includegraphics[width=0.3\linewidth]{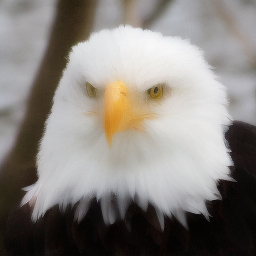}  & \includegraphics[width=0.3\linewidth]{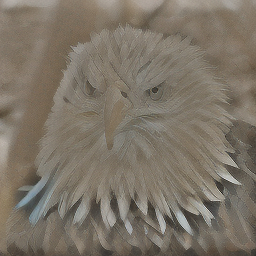} &\includegraphics[width=0.3\linewidth]{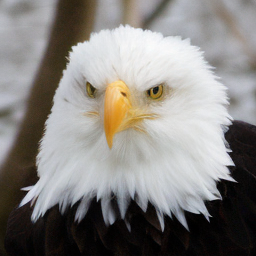} \\\includegraphics[width=0.3\linewidth]{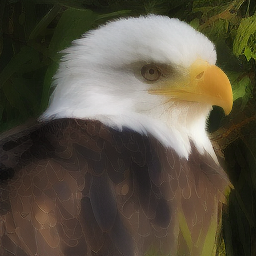} &\includegraphics[width=0.3\linewidth]{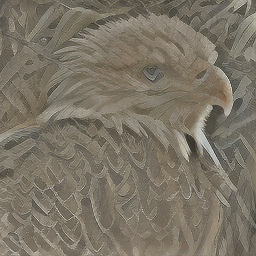} &\includegraphics[width=0.3\linewidth]{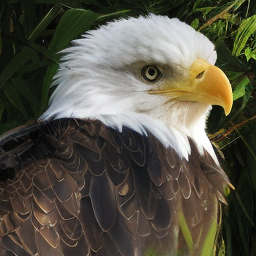}  \\
    \includegraphics[width=0.3\linewidth]{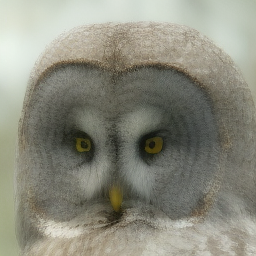}  & \includegraphics[width=0.3\linewidth]{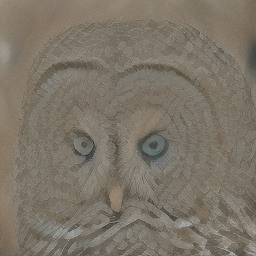} &\includegraphics[width=0.3\linewidth]{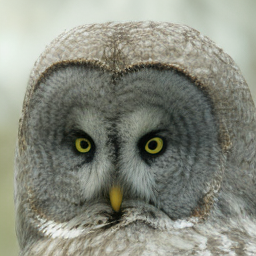} \\\includegraphics[width=0.3\linewidth]{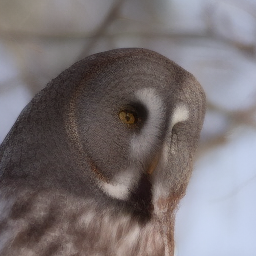} &\includegraphics[width=0.3\linewidth]{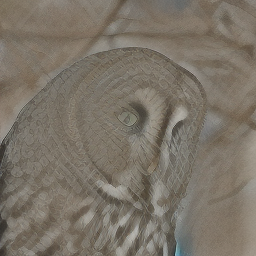} &\includegraphics[width=0.3\linewidth]{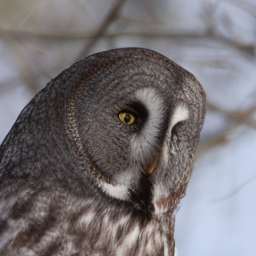}  \\
      Low-frequency Output & High-frequency Output & Final Output \\
    \end{tabular}
    \caption{
    \textbf{Visualization of generated low-, high-frequency and final outputs.} The final output from the spatial branch is enhanced by the low- and high-frequency information provided by the frequency branch.
    }
    \label{fig:freq1}
\end{figure*}

\begin{figure*}[t]
    \centering
    \setlength{\tabcolsep}{3pt}
    \begin{tabular}{ccc}
    \includegraphics[width=0.3\linewidth]{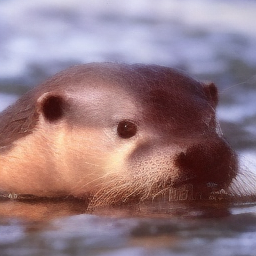}  & \includegraphics[width=0.3\linewidth]{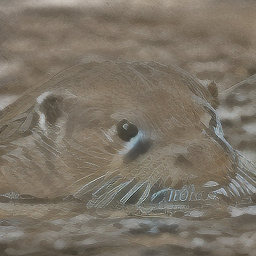} &\includegraphics[width=0.3\linewidth]{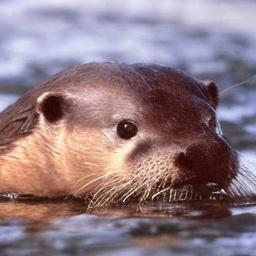} \\\includegraphics[width=0.3\linewidth]{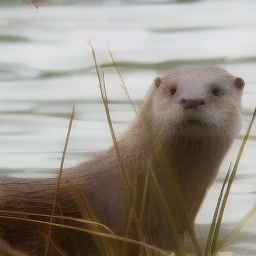} &\includegraphics[width=0.3\linewidth]{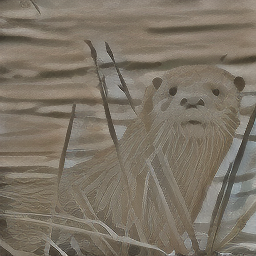} &\includegraphics[width=0.3\linewidth]{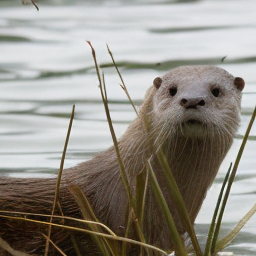}  \\
    \includegraphics[width=0.3\linewidth]{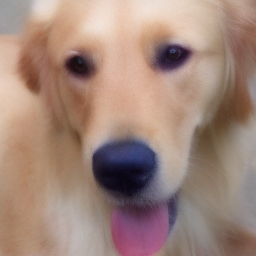}  & \includegraphics[width=0.3\linewidth]{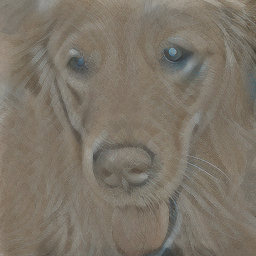} &\includegraphics[width=0.3\linewidth]{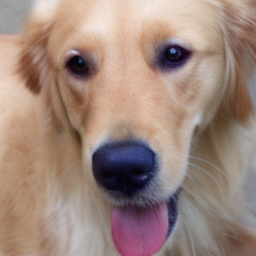} \\\includegraphics[width=0.3\linewidth]{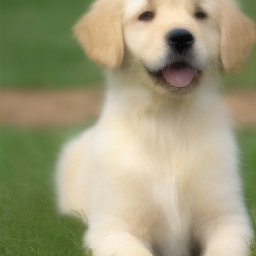} &\includegraphics[width=0.3\linewidth]{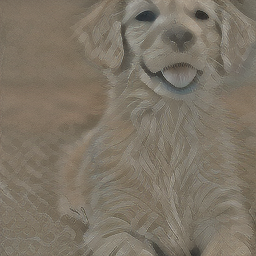} &\includegraphics[width=0.3\linewidth]{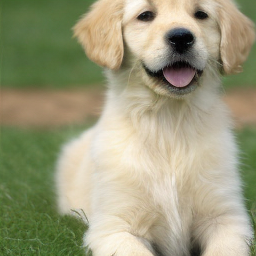}  \\
      Low-frequency Output & High-frequency Output & Final Output \\
    \end{tabular}
    \caption{
    \textbf{Visualization of generated low-, high-frequency and final outputs.} The final output from the spatial branch is enhanced by the low- and high-frequency information provided by the frequency branch.
    }
    \label{fig:freq2}
\end{figure*}

\begin{figure*}[t]
    \centering
    \setlength{\tabcolsep}{3pt}
    \begin{tabular}{ccc}
    \includegraphics[width=0.3\linewidth]{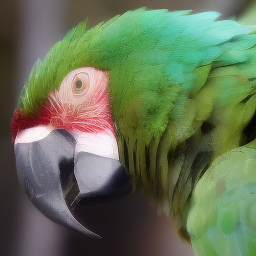}  & \includegraphics[width=0.3\linewidth]{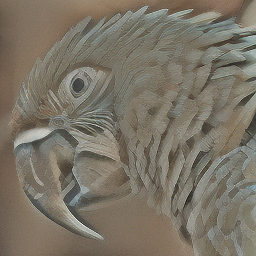} &\includegraphics[width=0.3\linewidth]{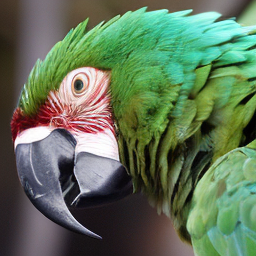} \\\includegraphics[width=0.3\linewidth]{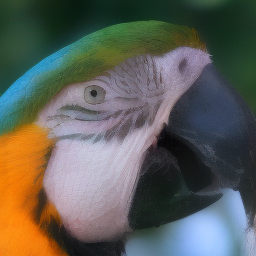} &\includegraphics[width=0.3\linewidth]{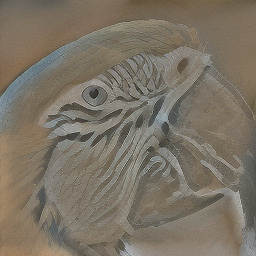} &\includegraphics[width=0.3\linewidth]{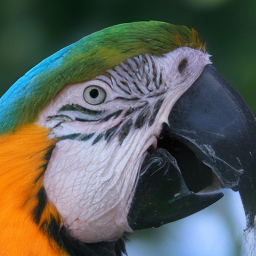}  \\
    \includegraphics[width=0.3\linewidth]{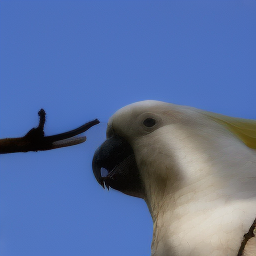}  & \includegraphics[width=0.3\linewidth]{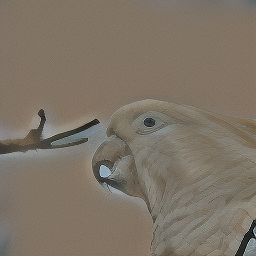} &\includegraphics[width=0.3\linewidth]{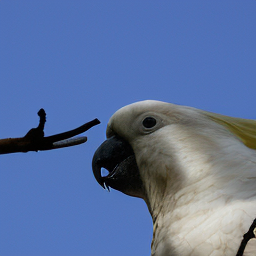} \\\includegraphics[width=0.3\linewidth]{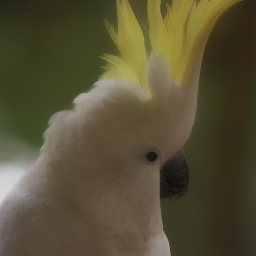} &\includegraphics[width=0.3\linewidth]{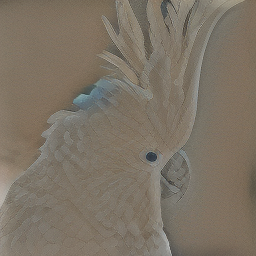} &\includegraphics[width=0.3\linewidth]{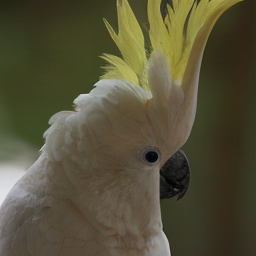}  \\
      Low-frequency Output & High-frequency Output & Final Output \\
    \end{tabular}
    \caption{
    \textbf{Visualization of generated low-, high-frequency and final outputs.} The final output from the spatial branch is enhanced by the low- and high-frequency information provided by the frequency branch.
    }
    \label{fig:freq3}
\end{figure*}

We provide additional visualizations of low- and high-frequency components (from the frequency branch) and the final output (from the spatial branch) in Fig.~\ref{fig:freq1} to Fig.~\ref{fig:freq3}.

\section{Visualization of Generated Samples}
\label{sec:sup_vis}
\begin{figure}
    \centering
    \includegraphics[width=\linewidth]{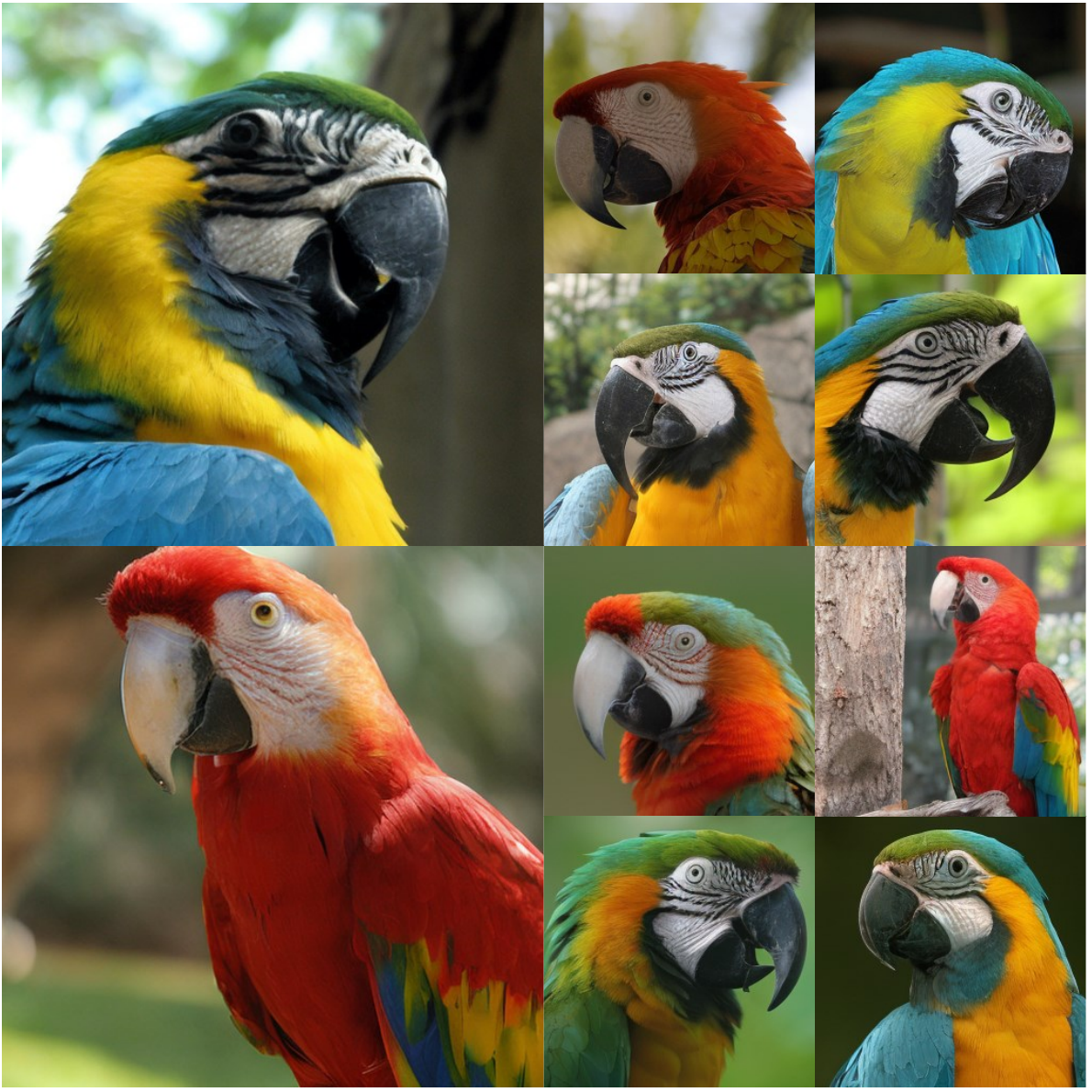}
    \caption{\textbf{Generated Samples from \modelname.} \modelname is able to generate high-quality golden retriever (88) images.}
    \label{fig:88}
\end{figure}

\begin{figure}
    \centering
    \includegraphics[width=\linewidth]{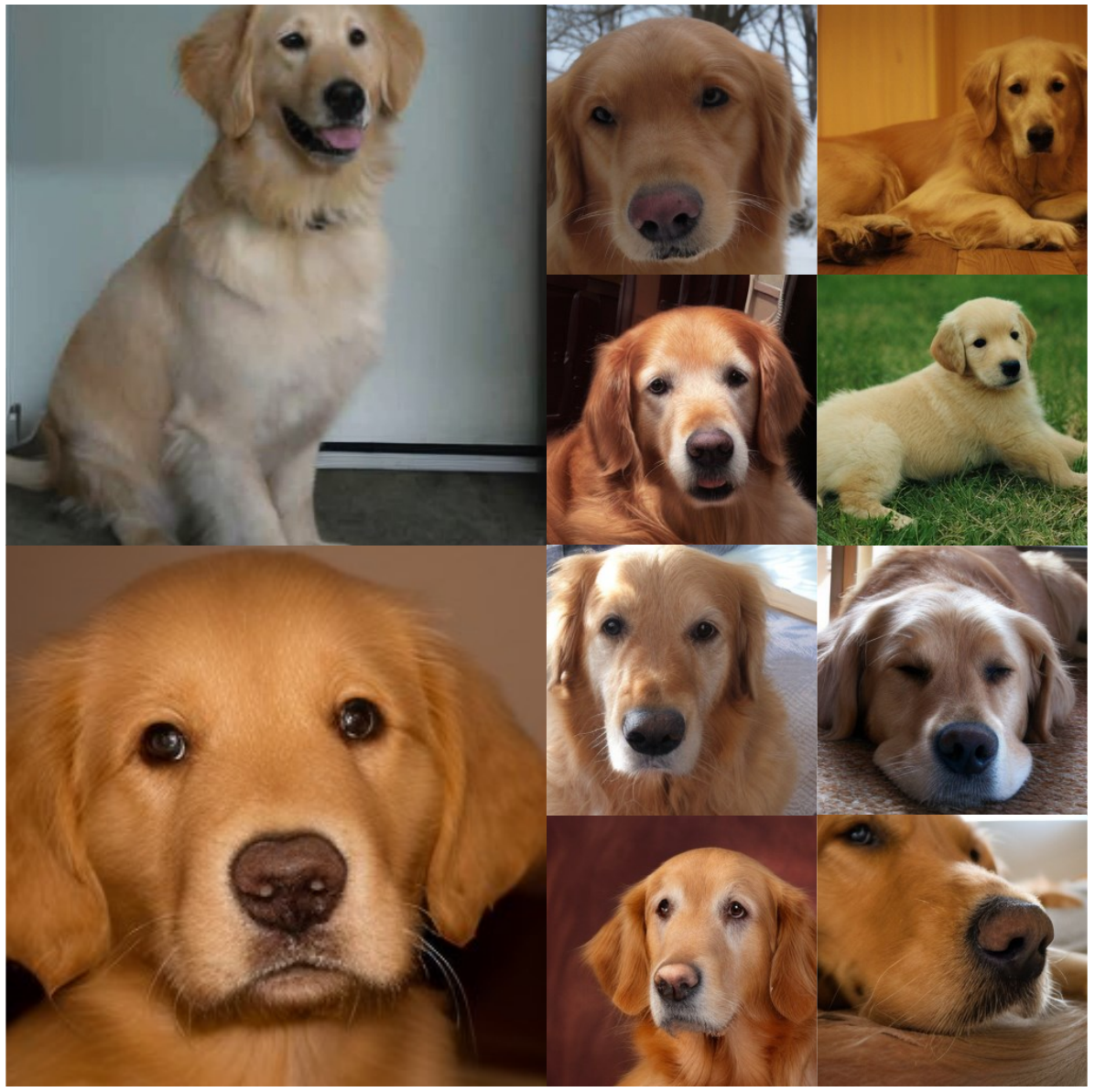}
    \caption{\textbf{Generated Samples from \modelname.} \modelname is able to generate high-quality golden retriever (207) images.}
    \label{fig:207}
\end{figure}

\begin{figure}
    \centering
    \includegraphics[width=\linewidth]{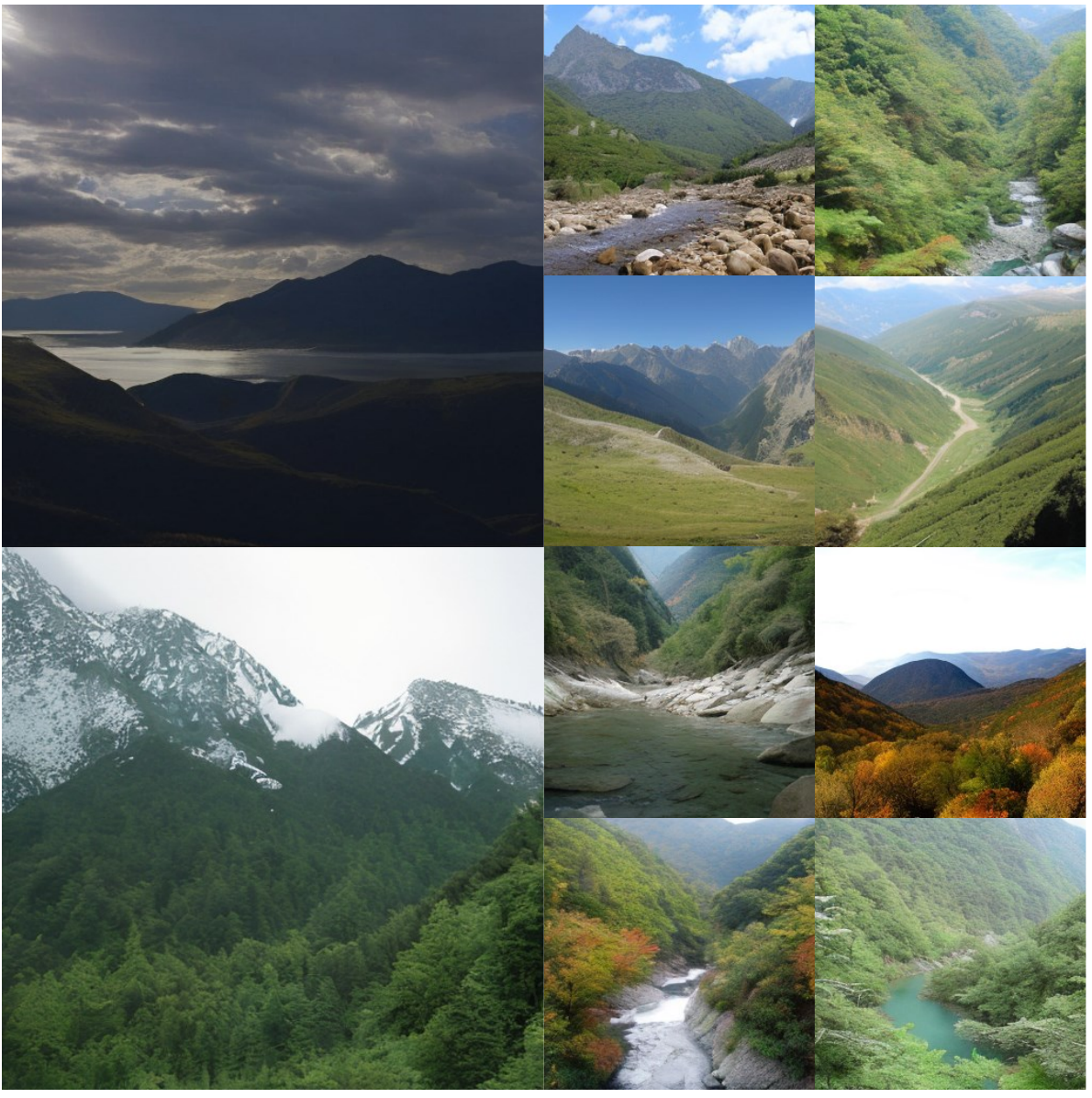}
    \caption{\textbf{Generated Samples from \modelname.} \modelname is able to generate high-quality golden retriever (979) images.}
    \label{fig:979}
\end{figure}

\begin{figure}
    \centering
    \includegraphics[width=\linewidth]{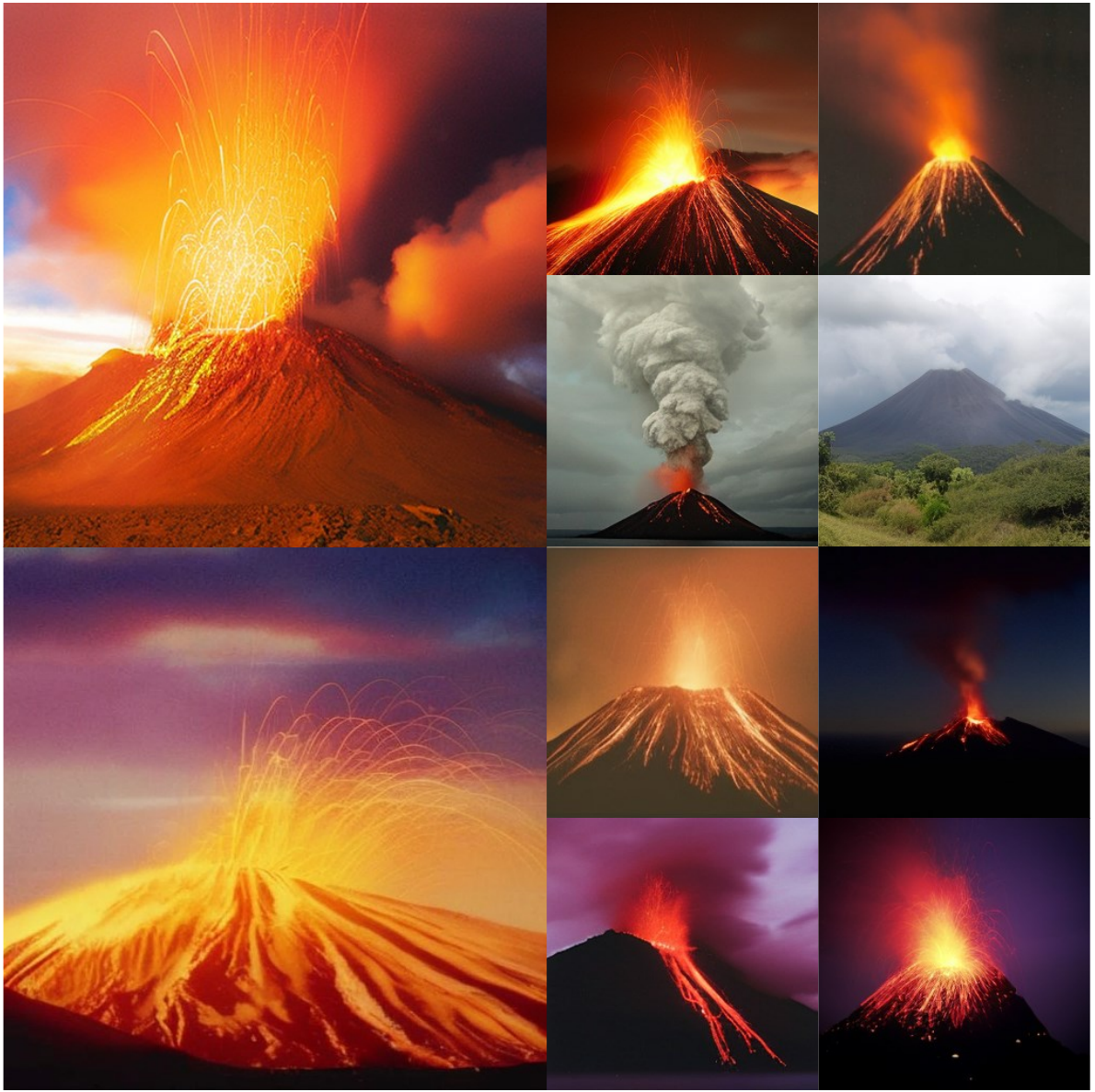}
    \caption{\textbf{Generated Samples from \modelname.} \modelname is able to generate high-quality golden retriever (980) images.}
    \label{fig:980}
\end{figure}

We provide additional visualization results from the spatial branch of \modelname in Fig.~\ref{fig:88} to Fig.~\ref{fig:980}.

\end{document}